\documentclass{article}
\usepackage{natbib}
\setcitestyle{numbers,square}
\usepackage{amsmath}
\usepackage{amsfonts} 
\usepackage{amsthm}
\usepackage{algorithm}
\usepackage{algorithmic}
\usepackage[preprint]{neurips_2023}
\usepackage{graphicx}
\usepackage{subfigure}
\usepackage[utf8]{inputenc} 
\usepackage[T1]{fontenc} 
\usepackage{hyperref} 
\usepackage{url} 
\usepackage{booktabs} 
\usepackage{nicefrac} 
\usepackage{microtype} 
\usepackage{xcolor} 
\usepackage{array,tabularx}

\title{A Novel Framework for Improving the Breakdown Point of Robust Regression Algorithms}

\author{
$\textbf{Zheyi Fan}^{1,2},\ \textbf{Szu Hui Ng}^{3}\ ,\ \textbf{Qingpei Hu}^{1,2}$\\
$^{1}$Academy of Mathematics and Systems Science, Chinese Academy of Sciences, China\\
$^{2}$School of Mathematical Sciences, University of Chinese Academy of Sciences, China\\
$^{3}$Department of Industrial Systems Engineering \& Management, National University of Singapore, Singapore\\
$^{1,2}$\texttt{\{fanzheyi,qingpeihu\}@amss.ac.cn}, $^{3}$\texttt{isensh@nus.edu.sg}\\
}

\begin{document}
\maketitle 
\begin{abstract}
We present an effective framework for improving the breakdown point of robust regression algorithms. Robust regression has attracted widespread attention due to the ubiquity of outliers, which significantly affect the estimation results. However, many existing robust least-squares regression algorithms suffer from a low breakdown point, as they become stuck around local optima when facing severe attacks. By expanding on the work of \cite{my_first}, we propose a novel framework that enhances the breakdown point of these algorithms by inserting a prior distribution in each iteration step, and adjusting the prior distribution according to historical information. We apply this framework to a specific algorithm and derive the consistent robust regression algorithm with iterative local search (CORALS). The relationship between CORALS and momentum gradient descent is described, and a detailed proof of the theoretical convergence of CORALS is presented. Finally, we demonstrate that the breakdown point of CORALS is indeed higher than that of the algorithm from which it is derived. We apply the proposed framework to other robust algorithms, and show that the improved algorithms achieve better results than the original algorithms, indicating the effectiveness of the proposed framework. 
\end{abstract}
\section{Introduction}
Robust regression is an important problem in machine learning, focusing on learning reliable information in the event of the pollution of, or attack on, a dataset. This technique has been applied in various fields to protect against abnormal events, such as computer vision~\cite{Robust_Regression_Methods_for_Computer_Vision,Lathuiliere_2018_ECCV}, biostatistics \cite{heritier2009robust}, and economics~\cite{hao2020forecasting}.

In this paper, we mainly focus on robust linear square regression (RLSR). In the RLSR problem, we are given a data matrix $X=[\mathbf{x}_{1},...,\mathbf{x}_{n}]\in \mathbb{R}^{d\times n}$, the corresponding response vector $\mathbf{y}\in \mathbb{R}^{n}$ (where $n, d$ represent the number of samples and the dimension of the data, respectively), and a non-negative integer $k$ indicating that there are $k$ corruptions in the response vector \textbf{y}. In general, the RLSR problem can be described as:
\begin{equation}
(\hat{\mathbf{w}},\hat S)=\arg\min_{\substack{\mathbf{w}\in\mathbb{R}^{p},S\subset[n]\\ |S|=n-k}}\sum_{i \in S}(y_{i}-\mathbf{x}_{i}^{T}\mathbf{w})^2 \label{problem}
\end{equation}
The purpose of RLSR is to discover the best point set $S$ on which the calculated regression coefficient $\mathbf{w}^*$ will lead to the minimum regression error. However, this problem is difficult to optimize directly as it is NP-hard \cite{Recovery_of_Sparsely_Corrupted_Signals}.

In general, the corruption of a dataset can be roughly divided into two categories: oblivious adversarial attack (OAA) and adaptive adversarial attack (AAA). Most proposed methods are devoted to maintaining a high breakdown point under these two attacks. The breakdown point $\alpha$ is a measure of robustness, representing the proportion of corruptions in the dataset that the RLSR algorithm can tolerate.

In the case of OAAs, where the opponent generates $k$ corruptions while completely ignoring $X$, $\mathbf{w}^*$, and $\boldsymbol{\epsilon}$, where $\boldsymbol{\epsilon}$ is the white noise in the model, there are several good solutions. Bhatia et al.~\cite{bhatia2017consistent} developed the first consistent estimator under mild conditions by using a hard thresholding operator, and Suggala et al.~\cite{pmlr-v99-suggala19a} extended their results to derive an excellent algorithm in which $\alpha$ gets close to 1 as $n \to \infty$. Using a different approach, Prasad et al.~\cite{prasad2018robust} implemented a novel robust variant of gradient descent that is robust for general statistical models, such as the classical Huber epsilon-contamination model, and in heavy-tailed settings.

Another active research area on robust regression has focused on handling the more challenging AAAs, in which opponents can view $X$, $\mathbf{w}^*$, and $\boldsymbol{\epsilon}$ before determining corruptions. Bakshi et al.~\cite{Optimal_Rates_in_Polynomial_Time} proposed an algorithm with optimal convergence speed based on the application of the SOS algorithm. In addition, SOS algorithm has also been extensively explored by Klivans et al.~\cite{klivans2018efficient}, Cherapanamjeri et al.~\cite{Cherapanamjeri2020}, and Zhu et al.~\cite{zhu2022robust} in robust regression. 
Diakonikolas et al.~\cite{pmlr-v97-diakonikolas19a,diakonikolas2019efficient} achieved robust estimation by using a kind of filter to wipe out some possible outliers in the iteration, and in \cite{diakonikolas2019efficient} they considered the situation in which both $X$ and $\mathbf{y}$ can be corrupted, eventually reaching an error bound of $\textit{O} (\alpha \log (1 / \alpha)\sigma) $. Bhatia et al.~\cite{bhatia2015robust} discovered a thresholding operator-based algorithm that searches for the best regression subset and produces consistent results under a noiseless model, i.e., $\boldsymbol{\epsilon} \equiv 0$ with a breakdown point of $1/65$.

Unlike traditional ideas that use different loss functions or optimization methods, another approach is to incorporate additional obtainable information to enhance the breakdown point of the robust regression problem in the case of AAAs. Fan et al.~\cite{my_first} reported that the breakdown point of some robust regression algorithms with highly nonconvex objective function could be significantly increased by incorporating a prior distribution $p_{\mathbf{w}}(\mathbf{w})$, at the cost of some bias in the final estimation. This inspires us to consider whether, if we can control the amplitude of this bias, we could eliminate it through an iterative framework. Thus, even if we do not have any prior knowledge of the parameter, the breakdown point of the original algorithm can be improved without creating any additional bias.

In this study, we extend the work of Fan et al.~\cite{my_first} and propose a new robust regression framework with an iteratively adjusted prior (REWRAP) that enhances the breakdown point for robust regression algorithms with highly nonconvex objective function. This framework will not lead to any additional bias, which is not availible in \cite{my_first}. This framework inserts a prior distribution of the parameter into the original robust regression algorithm in each iteration, where the prior is calculated using the result of the previous iteration. By applying this framework to the CRR robust regression algorithm \cite{bhatia2017consistent}, we derive a new and efficient consistent robust regression algorithm with iterative local search (CORALS). We provide a detailed proof of the theoretical convergence properties of CORALS and demonstrate that the theoretical breakdown point is indeed improved compared with that of the original CRR. We also investigate the relationship between CORALS and the traditional momentum gradient descent, providing evidence that momentum with past information may help the algorithm to jump away from local optima. We apply the framework to other algorithms, and implement extensive experiments to demonstrate the improvement in performance under OAAs and AAAs. The experimental results illustrate that our framework improves the breakdown point of the original algorithm under such attacks, which verifies that the framework effectively improves the robustness of different methods.

\textbf{Contribution:} 
The main contribution of this paper is the REWRAP framework, which effectively improves the breakdown point for some robust regression algorithms with strongly nonconvex objective function. We apply this framework to the CRR algorithm \cite{bhatia2017consistent} and derive the convergence properties of the resulting CORALS, which illustrates that we can always find a proper prior to increase the breakdown point. We also show that CORALS is similar to momentum gradient descent in some ways. Extensive experiments demonstrate that the REWRAP framework improves the breakdown point for various robust regression algorithms under OAAs and AAAs, which proves the effectiveness of our method.

\textbf{Paper Organization: }
In Section 2, we introduce the problem formulation and describe some notation and tools. We state the details of the proposed REWRAP framework and CORALS in Section 3. In Section 4, we list the theoretical properties of CORALS. Section 5 presents extensive experimental results on the parameter recovery effects of applying the REWRAP framework. Finally, Section 6 concludes this paper.

\section{Problem Formulation}
The aim of this work is to increase the breakdown point of RLSR algorithms under AAAs. In the RLSR setting, we are given a covariant matrix $X=[\mathbf{x}_{1},...,\mathbf{x}_{n}]\in \mathbb{R}^{d\times n}$, where $\mathbf{x}_{i} \in \mathbb{R}^{d}$. To represent the data generation in the event of data corruption, a commonly used model can be written as \[\mathbf{y}=X^{T}\mathbf{w}^{*}+\mathbf{b}^{*}+\boldsymbol{\epsilon} \]
where $\mathbf{w}^*$
is the true regression coefficient and $\boldsymbol{\epsilon}$ is a dense white noise vector subject to a specific distribution, which means that $\left\|\boldsymbol{\epsilon} \right\|_{0}\sim{n}$. The vector $\mathbf{b}^{*}$ is $k$-sparse, with only $k$ nonzero values, indicating $k$ unbounded noise terms in the response vector.

We propose a framework that enhances the breakdown point for a certain type of RLSR algorithm, which will be easily stuck into local optima because of the highly nonconvex objective function. Then we try to prove some theoretical properties of this framework through applying it on a specific RLSR algorithm, CRR. We will demonstrate that the breakdown point does indeed increase under some mild conditions. Following the work of Bhatia et al. \cite{bhatia2015robust}, we require two important properties in the convergence theory: \textit{Subset Strong Convexity (SSC)} and \textit{Subset Strong Smoothness (SSS)}. These two properties ensure that the distribution of the independent variable is not too abnormal, and should even hold for a subset of the data. This enables us to discover the convergence theory for any proportion of data. Given a set $S\subset [n]$, where $[n]={1,2,...,n}$, $X_{S}:=[\mathbf{x}_{i}]_{ i\in S}\in \mathbb{R}^{d\times |S|}$ signifies the matrix with columns in the set $S$. The minimum and maximum eigenvalues of a square matrix $X$ are denoted by $\lambda_{min}(X)$ and $\lambda_{max}(X)$, respectively.

\theoremstyle{plain} 
\newtheorem{definition}{Definition}
\begin{definition}[SSC Property]
A matrix $X \in \mathbb{R}^{d\times n}$ is said to satisfy the SSC property at level $m$ with constant $\lambda_{m}$ if the following holds:
\begin{equation}
\lambda_{m}\le \min_{|S|=m}\lambda_{min}(X_{S}X_{S}^{T})
\end{equation}
\end{definition}
\begin{definition}[SSS Property]
A matrix $X \in \mathbb{R}^{d\times n}$ is said to satisfy the SSS property at level $m$ with constant $\Lambda_{m}$ if the following holds:
\begin{equation}
\max_{|S|=m}\lambda_{max}(X_{S}X_{S}^{T}) \le \Lambda_{m}
\end{equation}
\end{definition}

\begin{algorithm}[tbp]
\caption{\textbf{REWRAP}: \textbf{R}obust r\textbf{E}gression frame\textbf{W}ork with ite\textbf{R}atively \textbf{A}djusted \textbf{P}rior}
\label{alg:REWRAP}
\begin{algorithmic}[1]
\REQUIRE Covariates $X=[\mathbf{x}_{1},...,\mathbf{x}_{n}]$, responses $\mathbf{y}=[y_{1},...,y_{n}]^T$, corruption index $k$, tolerance $\epsilon$,
\\
deviation robust estimate $\hat{\sigma}$
\ENSURE solution $\hat{\mathbf{w}}$
\STATE Initialize prior distribution coefficient $\mathbf{\theta}^0$, $t\gets 0$
\WHILE {not converged}
\STATE $p_{post}^{t+1}(\mathbf{w})$=Robust\_Regression\big($X,\mathbf{y},k,\hat{\sigma},p_{\mathbf{w}}(\mathbf{w}|\mathbf{\theta}^{t})$\big);
\STATE $p_{\mathbf{w}}(\mathbf{w}|\mathbf{\theta}^{t+1})$=Update$\left(p_{post}^{t+1}(\mathbf{w}),\mathbf{\theta}^{t}\right)$
\STATE $t\gets t+1$;
\ENDWHILE
\RETURN $\hat{\mathbf{w}} \gets MAP\left(p_{post}^{t}(\mathbf{w})\right)$
\end{algorithmic}
\end{algorithm}
\begin{algorithm}[tbp]
\caption{Simple Normal Prior Update Strategy}
\label{alg:SNP}
\begin{algorithmic}[1]
\REQUIRE Posterior distribution of parameters $\mathbf{w}$: $p_{post}(\mathbf{w})$, covariance matrix $\Sigma$
\ENSURE adjusted prior distribution $p_{\mathbf{w}}(\mathbf{w})$
\STATE set $\mu=MAP(p_{post}(\mathbf{w}))$
\RETURN $p_{\mathbf{w}}(\mathbf{w})\gets \mathcal{N}(\mathbf{w}| \mu, \Sigma)$
\end{algorithmic}
\end{algorithm}
The conditions for satisfying the SSC and SSS properties are provided in Appendix B. These two properties will be applied in the proof presented in Section \ref{Theoretical Analysis}.

\section{Methodology}
In this section, we first introduce our REWRAP framework in Section 3.1 to state the basic premise of how it improves the breakdown point of robust regression. We then utilize this framework on a special RLSR algorithm, CRR, and create a new algorithm, CORALS, in Section 3.2. The relationship between CORALS and momentum gradient descent is described in Section 3.3.

\subsection{Robust Regression Framework with Iteratively Adjusted Prior}

We propose the novel REWRAP framework to improve the breakdown point of RLSR algorithms. This framework is designed to utilize historical information, namely the posterior distribution of the parameter $p_{post}^{t+1}(\mathbf{w})$ calculated in the previous step, to adjust the coefficients in the prior distribution. This new prior is used in the next iteration as a constraint, forcing the algorithm to search for a solution around the prior mean. The details of the REWRAP framework can be seen in Algorithm \ref{alg:REWRAP}. The `Robust\_Regression' in Algorithm \ref{alg:REWRAP} refers to any robust regression algorithm that satisfies the following key property: the breakdown point increases when the prior distribution is incorporated, though this will create some bias in the final estimation, which will be fully discussed in Section \ref{Theoretical Analysis}. For example, both the TRIP and BRHT algorithms proposed in \cite{my_first} satisfy this key property. Thus, as long as we can control the amplitude of the bias in each iteration by setting a proper parameter in the prior distribution, the bias will continually decrease as the number of iterations grows. This should satisfy $\text{bias}^{t+1}\le \beta \text{bias}^{t}$, where $\beta<1$ is a constant and $\text{bias}^t$ is the estimation bias in the $t^{th}$ iteration. There will also be a certain amount of growth in the breakdown point of the algorithm due to the incorporated prior. 
\begin{algorithm}[tbp]
\caption{\textbf{CORALS}:\textbf{C}onsistent r\textbf{O}bust \textbf{R}egression with iter\textbf{A}tive loca\textbf{L} \textbf{S}earch}
\label{alg:CORALS}
\begin{algorithmic}[1]
\REQUIRE Covariates $X=[\mathbf{x}_{1},...,\mathbf{x}_{n}]$, responses $\mathbf{y}=[y_{1},...,y_{n}]^T$,
penalty matrix $M$,\\
corruption index $k$, tolerance $\epsilon$
\ENSURE solution $\hat{\mathbf{w}}$
\STATE Initialize $\mathbf{w}^0=(XX^{T})^{-1}(X\mathbf{y})$, $t\gets 0$
\WHILE{$\|\mathbf{w}^{t}-\mathbf{w}^{t-1}\|_{2}>\epsilon$}
\STATE $\mathbf{w}^{t+1} \gets TRIP(X,\mathbf{y}, \mathbf{w}^t,M,k,\epsilon)$
\STATE $t\gets t+1$
\ENDWHILE
\RETURN $\hat{\mathbf{w}} \gets \mathbf{w}^{t}$
\end{algorithmic}
\end{algorithm}

\begin{algorithm}[tbp]
\caption{\textbf{TRIP}: hard \textbf{T}hresholding approach to \textbf{R}obust regression with s\textbf{I}mple \textbf{P}rior}
\label{alg:TRIP}
\begin{algorithmic}[1]
\REQUIRE Covariates $X=[\mathbf{x}_{1},...,\mathbf{x}_{n}]$, responses $\mathbf{y}=[y_{1},...,y_{n}]^T$,
prior knowledge $\mathbf{w}_0$, \\penalty matrix $M$,
corruption index $k$, tolerance $\epsilon$
\ENSURE solution $\hat{\mathbf{w}}$
\STATE $\mathbf{b}^0 \gets \mathbf{0}$, $s\gets 0$,\\
$P_{MX}\gets X^T(XX^T+M)^{-1}X$, $P_{MM}\gets X^T(XX^T+M)^{-1}M$
\WHILE {$\|\mathbf{b}^{s}-\mathbf{b}^{s-1}\|_{2}>\epsilon$}
\STATE $\mathbf{b}^{s+1}\gets HT_k(P_{MX} \mathbf{b}^{s}+(I-P_{MX})\mathbf{y}-P_{MM}\mathbf{w}_0)$
\STATE $s\gets s+1$;
\ENDWHILE
\RETURN $\hat{\mathbf{w}} \gets (XX^T)^{-1}X(\mathbf{y}-\mathbf{b}^{s})$
\end{algorithmic}
\end{algorithm}
In this work, we apply a simple but efficient strategy in the prior update. We set the prior distribution of $\mathbf{w}$ in the form of a normal distribution $\mathcal{N} (\mathbf{w}|\mu_{t},\Sigma_{t})$ in the $t^{th}$ step of REWRAP. We choose $\Sigma_{t}$ to be a constant matrix $\Sigma$ that does not vary between iterations. $\mu_{t}$ is selected to be the maximum a posteriori (MAP) estimation of the posterior distribution $p_{post}^{t+1}(\mathbf{w})$ in the previous step, because MAP estimation is much easier than other methods as it does not require the full posterior distribution. We do not use $\Sigma_{t}$ as a prior and estimate it based on the posterior distribution because, if the estimation error in the initial iterations is large, the posterior distribution will have a large variance. This will decrease the effect of the prior in the estimation of $\mathbf{w}$, which will lead to the algorithm becoming stuck around local optima. This situation is avoided by using a constant matrix. The form of $\Sigma$ can be set as $\tau I$, where the coefficient $\tau$ is a positive number selected by 5-fold or 10-fold cross-validation. The experimental results in Section \ref{Experiments} show that REWRAP improves many traditional algorithms, demonstrating the effectiveness of our framework.

\subsection{Consistent Robust Regression with Iterative Local Search}

We now describe the application of REWRAP to a specific RLSR algorithm, CRR \cite{bhatia2017consistent}, and propose a new robust regression algorithm, CORALS. The process of CORALS is given in Algorithm \ref{alg:CORALS}. Details of incorporating a prior into CRR can be found in the TRIP algorithm~\cite{my_first}, which is also a sub-algorithm in CORALS, as seen in Algorithm \ref{alg:TRIP}. The hard thresholding operator $HT(\cdot)$ in the TRIP algorithm is defined as follows.
\begin{definition}[Hard Thresholding]
For any vector $\mathbf{r} \in \mathbb{R}^n$, let $\delta_{\mathbf{r}}^{-1}(i)$ represent the position of the $i^{th}$ element in $\mathbf{r}$, where the elements are arranged in descending order of magnitude. Then, for any $k<n$, the hard thresholding operator is defined as $\hat{\mathbf{r}}=HT_k(\mathbf{r})$, where $\hat{\mathbf{r}}_i=\mathbf{r}_i$ if $\delta_{\mathbf{r}}^{-1}(i)\le k$ and $0$ otherwise.
\end{definition} 
With a prior distribution $p_{\mathbf{w}}(\mathbf{w})=\mathcal{N} (\mathbf{w}^t,\Sigma)$, the TRIP algorithm attempts to solve the original optimization problem in Eq. (\ref{problem}) with an additional quadratic penalty term:
\begin{equation}
(\hat{\mathbf{w}},\hat S)=\arg\min_{\substack{\mathbf{w}\in\mathbb{R}^{p},S\subset[n]\\ |S|=n-k}}\sum_{i \in S}(y_{i}-x_{i}^{T}\mathbf{w})^2 +(\mathbf{w}-\mathbf{w}^t)^{T}M(\mathbf{w}-\mathbf{w}^t)\label{problem simple}
\end{equation}
where $M=(\Sigma/\sigma^2)^{-1}$. CORALS is actually an iterative TRIP algorithm with a continuously updated prior, which effectively searches the solution space around the last estimation. There is a simple relationship between the original CRR and CORALS. Suppose the penalty matrix $M$ is in the form of $\tau I$. Then, if $\tau=0$ or $+\infty$, CORALS degenerates into the CRR algorithm, which means that CRR can be treated as a special case of CORALS.

This simple improvement enhances the breakdown point of the original CRR. In Section 5, we show that the estimation bias in each step of CORALS can be controlled. Therefore, CORALS converges under a weaker condition than CRR by incorporating a prior, and guarantees a consistent unbiased result. We also present a theoretical optimal penalty matrix $M$ in the form of $\tau I$.

\subsection{Relationship Between CORALS and Momentum Gradient Descent } 
In this subsection, we reveal the similarity between CORALS and momentum gradient descent. In each iteration of CORALS, we implement the TRIP algorithm, which means we need to solve Eq. (\ref{problem simple}) iteratively. By incorporating a sparse corruption vector $\mathbf{b}$, Eq. (\ref{problem simple}) can be formulated as: 
\begin{equation} \label{TRIP 1}
\min_{\mathbf{w}\in\mathbb{R}^{d},\|\mathbf{b}\|_{0}\le k^*}
\frac{1}{2}\|X^T\mathbf{w}-(\mathbf{y}-\mathbf{b})\|_2^2 +\frac{1}{2}(\mathbf{w}-\mathbf{w}^t)^TM(\mathbf{w}-\mathbf{w}^t) 
\end{equation}
For any estimation $\hat{\mathbf{b}}$ of the corruption vector $\mathbf{b}^*$, a closed-form estimation of $\mathbf{w}^*$ can be easily calculated as $\hat{\mathbf{w}}=(XX^T+M)^{-1}[X(\mathbf{y}-\hat{\mathbf{b}})+M\mathbf{w}^t]$. By inserting this estimation into the optimization problem of Eq. (\ref{TRIP 1}), a clearer form of the objective function of TRIP is obtained:
\begin{equation}
\min_{\|\mathbf{b}\|_{0}\le k^*}
f_{corals}(\mathbf{b})=\frac{1}{2}\|(P_{MX}-I)(\mathbf{y}-\mathbf{b})+P_{MM}\mathbf{w}^t\|_2^2
\end{equation}
In this way, the estimation of $\mathbf{b}^{s}$ in TRIP during the $t^{th}$ iteration of CORALS can be expressed as $\mathbf{b}^{s+1}=HT_k(\mathbf{b}^{s}-\nabla f_{corals}(\mathbf{b}^{s}))$. Compared with the original CRR algorithm,
the objective function can be rewritten as: 
\begin{equation}
\min_{\|\mathbf{b}\|_{0}\le k^*}
f_{crr}(\mathbf{b})=\frac{1}{2}\|(P_{X}-I)(\mathbf{y}-\mathbf{b})\|_2^2
\end{equation}

where $P_X=X^{T}(XX^{T})^{-1}X$. Note that $\mathbf{w}^{t}= (XX^T)^{-1}X(\mathbf{y}-\hat{\mathbf{b}}_{t})$, where $\hat{\mathbf{b}}_{t}$ is the final corruption vector estimated by CORALS in the $t-1^{th}$ iteration. Through the above symbol definition and analysis process, we obtain the following relationship when $n$ is sufficiently large compared with $d$ and $M$ has the form $\tau I$.

\theoremstyle{plain} 
\newtheorem{theorem}{Theorem}
\setcounter{theorem}{0}
\renewcommand{\thetheorem}{\arabic{theorem}}
\begin{theorem}
Suppose that the number of samples $n$ and data dimension $d$ satisfy $d/n\to 0$, $M=\tau I$, and assume that $\mathbf{x}_i\in \mathbb{R}^d$ are generated from the standard normal distribution. Then, in the $t^{th}$ iteration of CORALS:
\[
\mathbf{b}^{s+1}= HT_k(P_{MX} \mathbf{b}^{s}+(I-P_{MX})\mathbf{y}-P_{MM}\mathbf{w}^{t})
\]
can be formally considered as:
\begin{align*}
\mathbf{b}^{s+1}
&=HT_k\left(\mathbf{b}^{s}-\nabla f_{corals}(\mathbf{b}^{s})\right)
\\
&=HT_k\left[\mathbf{b}^{s}-(A\nabla f_{crr}(\mathbf{b}^{s})+B\nabla f_{crr}(\hat{\mathbf{b}}_{t})+C)\right]
\end{align*}
where $A+B=I$, $\|C\|_2=O(\sqrt{d})$. 
\end{theorem}

This shows that CORALS uses similar ideas to momentum gradient descent (though the mixture of gradients is different from the original method). The application of this momentum idea makes it easier for the algorithm to jump away from local optima in Eq.~\ref{problem}, resulting in better results when facing data corruption.


\section{Theoretical Analysis}
\label{Theoretical Analysis}
In this section, we present the properties and theoretical results of our algorithms and show how they allow REWRAP to improve the breakdown point. We first examine the convergence of CORALS to demonstrate that this framework indeed increases the breakdown point of the original CRR algorithm. We then investigate the theoretical conditions that must be satisfied for REWRAP. 

We set
$\tilde{\mathbf{b}}=HT_{k}(\mathbf{b}^{*}+\boldsymbol{\epsilon})$, and in the $t^{th}$ step of CORALS, we define $I_{s}:=supp(\mathbf{b}^s)\cup supp(\tilde{\mathbf{b}})$, $\mathbf{g}=P_{MM}(\mathbf{w}^{*}-\mathbf{w}^{t})$. Then, we have the following results.
\setcounter{theorem}{1}
\renewcommand{\thetheorem}{\arabic{theorem}}
\begin{theorem}\label{theroem_2}
Let $X=[\mathbf{x}_1,\dots,\mathbf{x}_n]\in \mathbb{R}^{d\times n}$ be the given data matrix and $\mathbf{y}=X^{T}\mathbf{w}^{*}+\mathbf{b}^{*}+\boldsymbol{\epsilon}$ be the corrupted output with sparse corruptions of $\|\mathbf{b}^*\|_0\le k\cdot n$. The elements of $\boldsymbol{\epsilon}$ are independent and obey the normal distribution $\mathcal{N}(0,\sigma^2)$. For a specific positive semi-definite matrix $M$, the data matrix $X$ satisfies the SSC and SSS properties such that $2\frac{\Lambda_{2k}}{\lambda_{min}(XX^T+M)}<1$. Then, in the $t^{th}$ step of CORALS, if $k>k^{*}$, it is guaranteed with a probability of at least $1-\delta$ that, for any $\epsilon,\delta>0$, after $S_{0}=\textit{O}(\log (\frac{\|\mathbf{b}^*\|_2}{\varepsilon}))$ iterations of TRIP:
\begin{align*}
\|\mathbf{b}^{s}-\tilde{\mathbf{b}}\|_2
\le&
\epsilon
+
O(\sigma\sqrt{d\log(d)})
+
2\frac{(\sqrt{\Lambda_{2k}}+\gamma\sqrt{\Lambda_{k}})\lambda_{max}(M)}{\lambda_{min}(XX^T+M)}\|\mathbf{w}^{*}-\mathbf{w}^{t}\|_2
\end{align*}
\end{theorem}

\begin{theorem}\label{theroem_3}
Under the conditions of Theorem 2 and assuming that $\mathbf{x}_i\in \mathbb{R}^d$ are generated from the standard normal distribution, for $k>k^{*}$, it is guaranteed with a probability of at least $1-\delta$ that, if:
\begin{equation*}
2\frac{\sqrt{\Lambda_n}(\sqrt{\Lambda_{2k}}+\gamma\sqrt{\Lambda_{k}})\lambda_{max}(M)}{\lambda_n\lambda_{min}(XX^T+M)}<1
\end{equation*}
then, for any $\epsilon,\delta>0$, after $T_{0}=O(log(\frac{\|\mathbf{w}^{0}-\mathbf{w}^{*}\|_2}{\epsilon}))$ iterations of CORALS, the current estimation coefficient $\mathbf{w}^{t}$ will satisfy:
\[\|\mathbf{w}^{t}-\mathbf{w}^*\|_2
\le
\epsilon
+
O(\sigma\sqrt{\frac{d}{n}\log \frac{d}{\delta}})
+
\frac{1}{\gamma}O(\sigma)
\]
\end{theorem}

where $\gamma$ is the amplification factor in Assumption \ref{assumption 1}, which reflects the convergence accuracy. Theorems \ref{theroem_2} and \ref{theroem_3} can be proved under some very mild assumptions (see Appendix C), which only ensure the identification of the corruption vector $\mathbf{b}^{*}$ and some basic convergence properties. These two theorems provide two key conditions for the convergence of CORALS, which can be used to calculate the breakdown point. For Theorem \ref{theroem_2}, CORALS requires $2\frac{\Lambda_{2k}}{\lambda_{min}(XX^T+M)}<1$, and the error reduction coefficient in Theorem \ref{theroem_3} should be less than 1. We attempt to estimate the breakdown point under the special condition $M=\tau I$, $\gamma$ is set to $1$, and assume that we know the true corruption number $k^*$, that is, $k=k^*$. In this situation, the breakdown point of CORALS $\alpha=k/n$ can be calculated as:
\begin{align*}
&\max_{\tau\in \mathbb{R}_{+}, \alpha\in [0,1]} \alpha
\\
\begin{split}
s.t.\quad
&\left \{
\begin{array}{ll}
2\frac{\Lambda_{2k}}{\lambda_{n}+\tau}<1\\
2\tau\frac{\sqrt{\Lambda_n}(\sqrt{\Lambda_{2k}}+\sqrt{\Lambda_{k}})}{\lambda_n(\lambda_n+\tau)}<1 
\end{array} 
\right. 
\end{split} 
\end{align*}

These two conditions can be split into two parts: convergence conditions for the embedded algorithm and the overall convergence condition. As the weight of the prior increases, i.e., with a higher $\tau$, the convergence conditions for the embedded algorithm become easier to satisfy. The overall convergence condition actually guarantee the bias in each step will decrease exponentially. By incorporating the prior, however, the overall convergence condition becomes a burden, although a sufficiently small $\tau$ will always satisfy the overall convergence condition. The original CRR algorithm can be viewed as a specific form of CORALS in which $\tau=0$ or $+\infty$, as mentioned in Section 3.2. Thus, the breakdown point of CRR can be expressed as:
\begin{align*}
&\max_{\alpha\in [0,1]} \alpha
\\
s.t.\quad &2\frac{\Lambda_{2k}}{\lambda_{n}}<1 \quad \text{or}\quad
2\frac{\sqrt{\Lambda_n}(\sqrt{\Lambda_{2k}}+\sqrt{\Lambda_{k}})}{\lambda_n}<1
\end{align*}
This demonstrates that the breakdown point of CORALS is definitely larger than that of CRR when $\tau$ is small. The breakdown point of CORALS can actually be improved to $1\%$, almost twice the $0.6\%$ of CRR.

\begin{theorem}
Suppose that $\mathbf{x}_i\in \mathbb{R}^d$ are generated from the standard normal distribution. For $k>k^{*}$, it is guaranteed with a probability of at least $1-\delta$ that, if $M$ is in the form of $\tau I$, and $\gamma=1$, the maximum breakdown point of CORALS can reach up to $1\%$ when $\tau=0.049n$.
\end{theorem}
This shows that REWRAP provides a useful tool for increasing the breakdown point. Overall, without loss of generality, we can assume the breakdown point of a specific robust regression algorithm should satisfy the following condition:
\begin{equation*}
f_{bp}(n,d,\sigma,k^{*},\Sigma)\le 1
\end{equation*}
where $f_{bp}(\cdot)$ is a function of all variables used in the regression, which is an index reflecting the robustness of the current regression, and $\Sigma$ is the covariance matrix of $X$. The REWRAP framework will be effective in the situation where the above convergence condition is weakened by incorporating a prior: 
\begin{equation*}
f_{bp}(n,d,\sigma,k^{*},\Sigma,p_{\mathbf{w}}(\mathbf{w}^{t}))\le f_{bp}(n,d,\sigma,k^{*},\Sigma)
\end{equation*}
However, this process will also lead to a bias $f_{bias}(n,d,\sigma,k^{*},\Sigma,p_{\mathbf{w}}(\mathbf{w}^{t}))$. As long as the bias satisfies $f_{bias}(n,d,\sigma,k^{*},\Sigma,p_{\mathbf{w}}(\mathbf{w}^{t+1})) \le \beta f_{bias}(n,d,\sigma,k^{*},\Sigma,p_{\mathbf{w}}(\mathbf{w}^{t}))$ by choosing proper prior coefficients, where $\beta<1$ is a constant non-negative value, REWRAP will enhance the breakdown point of the original algorithm. All details of the proof are listed in Appendix C.

\section{Experiments}\label{Experiments}
In this section, we present the results of numerical experiments to verify the performance of various algorithms under different dataset attacks. Both OAA and AAA are used to corrupt the dataset.

\subsection{Data and Metrics}
Similar to the experiments implemented in \cite{my_first}, we generate the experimental data in two steps. In the first step, we set the basic linear model $y_{i}=\mathbf{x}_{i}^T{w}^*+\epsilon_{i}$, where the true coefficient $\mathbf{w}^*$ is generated from a random norm vector $\mathcal{N} (0,I_{{d}})$. The covariant $\mathbf{x}_i$ is independent and identically distributed in $\mathcal{N} (0,I_{{d}})$ and $\epsilon_{i}$ is independent and identically distributed in $\mathcal{N} (0,\sigma^2)$. We set $\sigma=1$ in all experiments. The second step is to corrupt the data by applying two categories of attacks: OAA and AAA, as described in Section \ref{Corruption Method}. These attacks create $k^*$ corrupted responses in the whole dataset. All parameters are fixed in each experiment.

To evaluate the algorithm performance, we apply the standard $L_2$ error to measure the estimation error: $r_{\hat{\mathbf{w}}}=\|\hat{\mathbf{w}}-\mathbf{w}^*\|_2$. The convergence criterion of each algorithm is set as $\|\mathbf{w}^{t+1}-\mathbf{w}^t\|_2\le 10^{-4}$. All results are averaged over 20 runs.
\subsection{Corruption Methods}\label{Corruption Method} 
In this subsection, we introduce the two attack types used in the experiments: OAA and AAA. The details of these two attacks are as follows.

\textbf{ OAA}: The set of corrupted points $S$ is selected as a uniformly random $k^{*}$-sized subset of $[n]$, and the corresponding response variables are set as $y_i=20+u_{i}$, where $u_i$ are sampled from the uniform distribution $U[0,15]$.

\textbf{ AAA}:
In the AAA setting, we use a kind of leverage-point attack to test the robustness of our methods. We set the attack as follows: choose $k^{*}$ points with the largest covariant norm $\|\mathbf{x}_i\|_2$ and set their corresponding $y_i$ to $0$, as the regression result will be strongly affected by high leverage points \cite{chatterjee1986influential}. Thus, the estimation will be more likely to have a large bias if the high leverage points are corrupted.
\begin{figure*}[tbp]
\centering
\subfigure[]
{
\begin{minipage}{3cm}
\centering
\includegraphics[width=3cm]{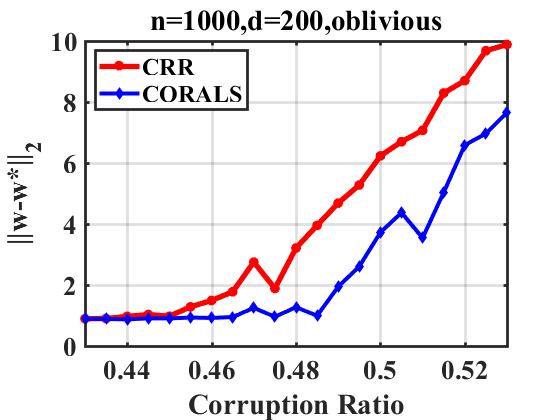}
\end{minipage}
}
\subfigure[]
{
\begin{minipage}{3cm}
\centering
\includegraphics[width=3cm]{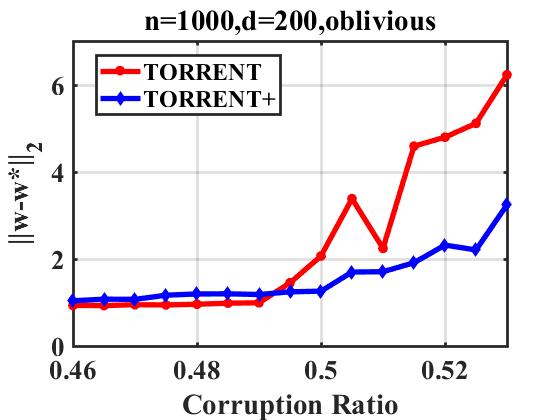}
\end{minipage}
}
\subfigure[]
{
\begin{minipage}{3cm}
\centering
\includegraphics[width=3cm]{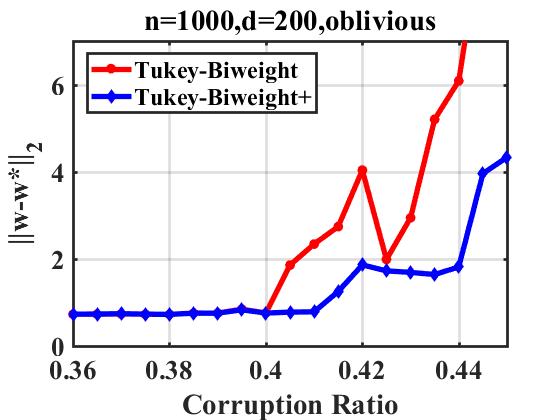}
\end{minipage}
}
\subfigure[]
{
\begin{minipage}{3cm}
\centering
\includegraphics[width=3cm]{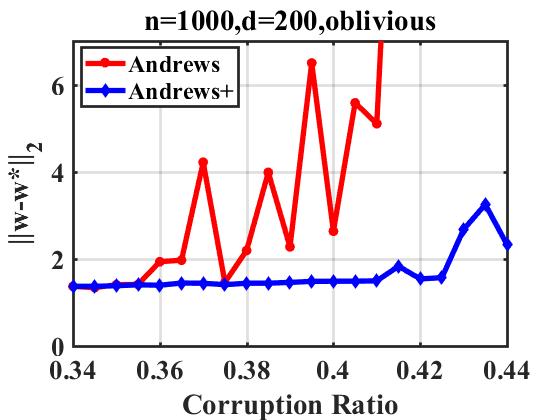}
\end{minipage}
}
\caption{Estimation error with respect to the number of data points $n$, dimension $d$, and corruption ratio $\alpha$ under OAA. All four regression algorithms are more robust under the REWRAP framework, with a greatly improved breakdown point.} 
\label{figure1}
\end{figure*}

\begin{figure*}[tbp]
\centering
\subfigure[]
{
\begin{minipage}{3cm}
\centering
\includegraphics[width=3cm]{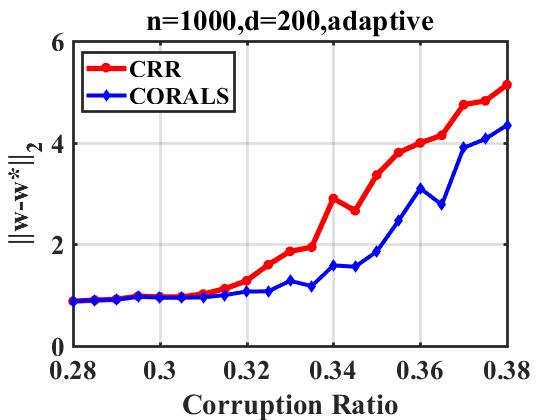}
\end{minipage}
}
\subfigure[]
{
\begin{minipage}{3cm}
\centering
\includegraphics[width=3cm]{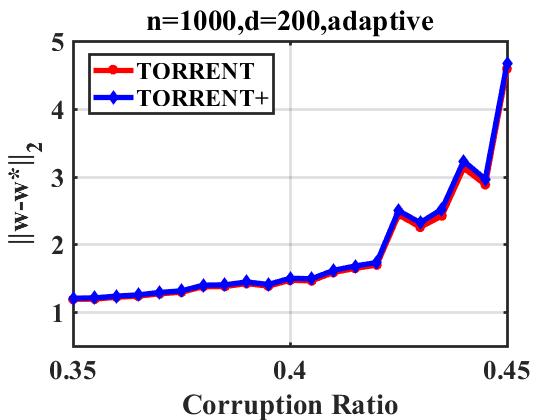}
\end{minipage}
}
\subfigure[]
{
\begin{minipage}{3cm}
\centering
\includegraphics[width=3cm]{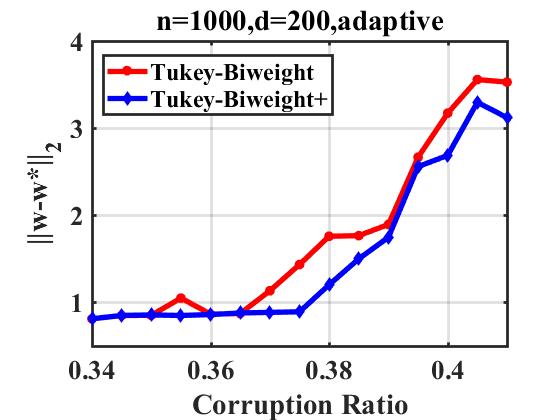}
\end{minipage}
}
\subfigure[]
{
\begin{minipage}{3cm}
\centering
\includegraphics[width=3cm]{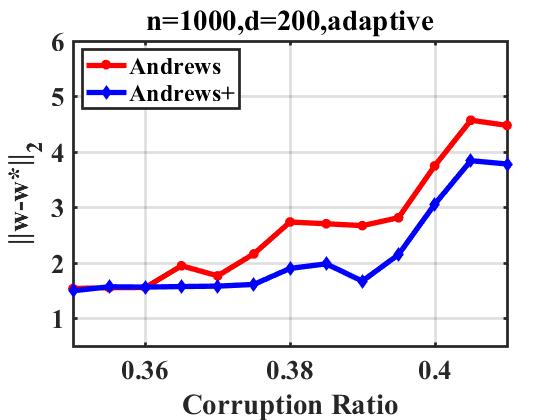}
\end{minipage}
}
\caption{Estimation error with respect to the number of data points $n$, dimension $d$, and corruption ratio $\alpha$ under AAA. The recovery effect is relatively poor compared with that under OAA, as the AAA situation is more complex to solve. In this case, however, there is still an improvement in the breakdown point, demonstrating the significance of the framework. } 
\label{figure2}
\end{figure*}
\subsection{Baseline Algorithms}
We test our iterative prior robust regression framework on four different algorithms, including two recent methods and two traditional M estimators. The two recent methods are the CRR algorithm \cite{bhatia2017consistent} and the TORRENT algorithm \cite{bhatia2015robust}, which perform well on RLSR problems. The two M estimators are the Tukey--Biweight estimator \cite{rey2012introduction} and the Andrews estimator \cite{rey2012introduction}, which have strongly nonconvex loss functions and easily become stuck around local optima when using normal optimization methods. The M estimator reaches a more robust result by using a generalized likelihood function of the form:
\begin{equation}
\max_{\mathbf{w}} \sum_{i=1}^{n}\rho\left(\frac{y_{i}-\mathbf{x}_{i}^{T}\mathbf{w}}{\sigma}\right)
\end{equation}
For the Tukey--Biweight and Andrews estimators, $\rho(\cdot)$ has the form:
\begin{align*}
&\rho_{Tukey}(x)=\left \{ 
\begin{array}{cc}
\frac{c_{Bi}^{2}}{6}\left[1-\left(1-\frac{x^2}{c_{Bi}^{2}}\right)\right], &|x|\le c_{Bi} \\
\frac{c_{Bi}^{2}}{6}, &|x|> c_{Bi} 
\end{array}
\right.
\\
&\rho_{Andrew}(x)=\left \{ 
\begin{array}{cc}
c_{An}^{2}\left[1-cos\left(\frac{x}{c_{An}}\right)\right], &|x|\leq \pi c_{An}\\
2c_{An}^{2}, &|x|> \pi c_{An}
\end{array}
\right.,
\end{align*}
and the coefficients of $\rho(\cdot)$ are $c_{Bi}=4.6851$ and $c_{An}=1.338$. We add a ``+" after the algorithm name to indicate that REWRAP has been applied, e.g., ``TORRENT+" and ``Tukey--Biweight+." CORALS is an exception, as it has been extensively used in the theoretical analysis. We set $\Sigma$ in the prior distribution as $\Sigma^{-1}=\tau I$ in all experiments. The method of inserting the prior distribution into the above algorithms and the prior coefficients under different attacks can be found in Appendix A.

\subsection{Recovery Effects}
Under OAAs, all four algorithms are much more robust when using the REWRAP framework than in their original form, as shown in Figure \ref{figure1}. With the REWRAP framework, the algorithms can tolerate a larger proportion of outliers without a significant increase in the error. For example, CORALS begins to become unstable when the corruption ratio exceeds $49\%$, while the original CRR can only tolerate $45\%$ outliers in the dataset, as seen in Figure \ref{figure1}(a). From Figure \ref{figure1}(d), Andrews+ can even tolerate $7\%$ more outliers than Andrews, which is a significant increase in the breakdown point. Even when both REWRAP-applied algorithm and the original begin to corrupt, the improved algorithm produces smaller estimation errors, indicating that it still maintains better robustness than the original algorithm.

This effect also appears in the AAA setting, but is a little weaker because AAAs are much more severe for robust regression tasks. CORALS, Tukey--Biweight+, and Andrews+ still behave better than the original algorithms, with the breakdown point improving by 1--2$\%$, and produce smaller estimation errors under higher corruption ratios. However, TORRENT+ behaves similarly to TORRENT, which is mainly because TORRENT is well-suited to this type of attack. When the corruption ratio exceeds $42\%$, the number of elements in the set $\{y_i|\, |y_i|<2\}$ will be more than $0.5n$, whereupon the algorithm is unable to distinguish between the true solution and the false parameter $\mathbf{w}_{false}=\mathbf{0}_d$. Thus, REWRAP cannot improve TORRENT under this level of AAA. In general, the REWRAP framework enhances the robustness of the original algorithms, leading to an increase in the breakdown point.

\section{Conclusion}
This paper has proposed a novel robust regression framework that enhances the robustness of regression algorithms. The REWRAP framework iteratively inserts a prior into the robust regression algorithm, and adjusts the prior through the results calculated in the previous iteration. We applied this framework to CRR to create CORALS as a demonstration of the framework's theoretical properties in this situation, and showed that the breakdown point of CRR could be almost doubled by using REWRAP. We also showed that CORALS has similarities with momentum gradient descent. Extensive experiments have demonstrated that the proposed framework significantly enhances the robustness of the original algorithm, illustrating the great value of REWRAP.

In this work, we only developed a very simple prior update strategy, which considers a nearly fixed normal distribution. Future research directions will include more complex prior distributions or more posterior information. Another research direction involves determining whether this framework can be applied to the case of covariate corruption.

\bibliography{reference_LP}

\newpage
\appendix
\theoremstyle{plain} 
\newtheorem{lemma}{Lemma}
\theoremstyle{plain} 
\newtheorem{theorem_apd}[lemma]{Theorem}
\theoremstyle{plain} 
\newtheorem{assumption}{Assumption}

\section{Method of Adding a Prior to the Model}
In this appendix, we discuss how to incorporate a prior distribution in a robust regression algorithm. We first begin with the M estimator to show the general method. Robust regression using the M estimator attempts to solve the following problem:
\begin{equation*}
\max_{\mathbf{w}} \sum_{i=1}^{n}\rho\left(\frac{y_{i}-\mathbf{x}_{i}^{T}\mathbf{w}}{\sigma}\right)
\end{equation*}
where $\sum_{i=1}^{n}\rho(\frac{y_{i}-\mathbf{x}_{i}^{T}\mathbf{w}}{\sigma})$ is a generalization of the log-likelihood. Thus, using traditional Bayesian statistics, if we have a prior distribution $p_{\mathbf{w}}(\mathbf{w})$, we can obtain the MAP estimate of $\mathbf{w}^{*}$ through the following optimization problem:

\begin{equation}\label{M_prior}
\max_{\mathbf{w}} \sum_{i=1}^{n}\rho\left(\frac{y_{i}-\mathbf{x}_{i}^{T}\mathbf{w}}{\sigma}\right)+\log p_{\mathbf{w}}(\mathbf{w})
\end{equation}

Then, if the prior distribution is in the form $\mathcal{N}(\mathbf{w}_0,\Sigma)$, Eq. (\ref{M_prior}) becomes:
\begin{equation*}
\max_{\mathbf{w}} \sum_{i=1}^{n}\rho\left(\frac{y_{i}-\mathbf{x}_{i}^{T}\mathbf{w}}{\sigma}\right)+(\mathbf{w}-\mathbf{w}_0)^{T}\Sigma^{-1}(\mathbf{w}-\mathbf{w}_0)
\end{equation*}
This problem can be solved by the iteratively reweighted least-squares (IRLS) algorithm \cite{aftab2015convergence}.

We now consider TORRENT \cite{bhatia2015robust}, which uses the following important definition:
\begin{definition}
For any vector $\mathbf{v}\in \mathbb{R}^{n}$, let $\sigma_{\mathbf{v}}\in S_{n}$ be the permutation that orders the elements of $\mathbf{v}$ in ascending order of magnitude, i.e., $|\mathbf{v}_{\sigma_{\mathbf{v}}(1)}|\le|\mathbf{v}_{\sigma_{\mathbf{v}}(2)}|\le...\le|\mathbf{v}_{\sigma_{\mathbf{v}}(n)}|$. Then, for any $k\le n$, we define the hard thresholding operator as:
\begin{equation*}
HTT(\mathbf{v},k)=\{i\in[n]:\sigma_{\mathbf{v}}^{-1}(i)\le k\}
\end{equation*}
\end{definition}
The $t^{th}$ step in TORRENT can then be expressed as:
\begin{align*}
&\text{Estimate\;} \mathbf{w}^{*}:\mathbf{w}^{t+1}=\arg\min_{\mathbf{w}}\sum_{i\in S_t}(y_{i}-\mathbf{x}_i^{T}\mathbf{w})^2
\\
&\text{Calculate\;} \mathbf{r}:\mathbf{r}^{t+1}=\mathbf{y}-X\mathbf{w}^{t}
\\
&\text{Estimate\;} S_{*}:S_{t+1}=HTT(\mathbf{r}^{t+1},(1-\beta n))
\end{align*}
where $\beta$ is a constant non-negative value. Thus, if we have a prior $p_{\mathbf{w}}(\mathbf{w})=\mathcal{N}(\mathbf{w}_0,\Sigma)$, we can add this to the `$\text{Estimate\;} \mathbf{w}^{*}$' step. Similar to the analysis of TRIP \cite{my_first}, this step can be transformed into the following form:
\begin{equation*}
\text{Estimate\;} \mathbf{w}^{*}:\mathbf{w}^{t+1}=\arg\min_{\mathbf{w}}\sum_{i\in S_t}(y_{i}-\mathbf{x}_i^{T}\mathbf{w})^2+(\mathbf{w}-\mathbf{w}_{0})^{T}M(\mathbf{w}-\mathbf{w}_{0})
\end{equation*}
where $M=(\Sigma/\sigma^2)^{-1}$. The full TORRENT+ pseudocode is given in Algorithm \ref{alg:TORRENT+}. In addition, the prior coefficients $\Sigma^{-1}=\tau I$ in Section \ref{Experiments} are listed in Tables \ref{tabel1} and \ref{tabel2}.

\begin{table}[h]
\centering
\begin{tabularx}{14em}%
{|*{2}{>{\centering\arraybackslash}X|}}
\hline
Algorithm & $\tau$ \\ \hline
CORALS & 0.049n \\ \hline
TORRENT+ & 0.01n \\ \hline
Tukey--Biweight+ & 0.002n \\ \hline
Andrews & 0.0035n \\ \hline
\end{tabularx}
\caption{Coefficients in the OAA setting} 
\label{tabel1}
\end{table}

\begin{table}[h]
\centering
\begin{tabularx}{14em}%
{|*{2}{>{\centering\arraybackslash}X|}}
\hline
Algorithm & $\tau$ \\ \hline
CORALS & 0.049n \\ \hline
TORRENT+ & 0.0001n \\ \hline
Tukey--Biweight+ & 0.0025n \\ \hline
Andrews & 0.0008n \\ \hline
\end{tabularx}
\caption{Coefficients in the AAA setting} 
\label{tabel2}
\end{table}

\begin{algorithm}[tbp]
\caption{TORRENT+}
\label{alg:TORRENT+}
\begin{algorithmic}[1]
\REQUIRE Covariates $X=[\mathbf{x}_{1},...,\mathbf{x}_{n}]$, responses $\mathbf{y}=[y_{1},...,y_{n}]^T$,
penalty matrix $M$,\\
threshold parameter $\beta$, tolerance $\epsilon$
\ENSURE solution $\hat{\mathbf{w}}$
\STATE Initialize $\mathbf{w}_0=(XX^{T})^{-1}X\mathbf{y}$, $t\gets 0$ 
\WHILE{$\|\mathbf{w}_{t}-\mathbf{w}_{t-1}\|_{2}>\epsilon$}
\STATE $S^{0}=[n]$, $\mathbf{r}^{0}=\mathbf{y}-X\mathbf{w}_{t}$, $s\gets 0$
\WHILE{$\|\mathbf{r}_{S_s}^{s}\|>\epsilon$}
\STATE $\mathbf{w}^{s+1}\gets \arg\min_{\mathbf{w}}\sum_{i\in S_s}(y_{i}-\mathbf{x}_i^{T}\mathbf{w})^2+(\mathbf{w}-\mathbf{w}_{t})^{T}M(\mathbf{w}-\mathbf{w}_{t})$\\
\STATE $\mathbf{r}^{s+1}\gets \mathbf{y}-X\mathbf{w}^{s}$
\STATE $S_{s+1}\gets HTT(\mathbf{r}^{s+1},(1-\beta)n)$
\STATE $s\gets s+1$
\ENDWHILE
\STATE $\mathbf{w}_{t+1} \gets \mathbf{w}^s$
\STATE $t\gets t+1$
\ENDWHILE
\RETURN $\hat{\mathbf{w}} \gets \mathbf{w}_{t}$
\end{algorithmic}
\end{algorithm}

\section{SSC/SSS Guarantees }
\setcounter{lemma}{7}
\renewcommand{\thelemma}{\arabic{lemma}}
In this section, we introduce some theoretical properties of SSC and SSS from \cite{bhatia2015robust}. These properties are used for the convergence analysis of the proposed algorithms. 

\begin{definition}
A random variable $x\in\mathbb{R}$ is said to be sub-Gaussian if the following quantity is finite:
\begin{equation*}
\sup_{p\ge 1}p^{-1/2}(E[|x|^p])^{1/p}
\end{equation*}
Moreover, the smallest upper bound on this quantity is referred to as the sub-Gaussian norm of $x$ and denoted as $\|x\|_{\psi_2}$.
\end{definition}

\begin{definition}
A vector-valued random variable $\mathbf{x}\in\mathbb{R}^d$ is said to be sub-Gaussian if its unidimensional marginals $\langle\mathbf{x},\mathbf{v}\rangle$ are sub-Gaussian for all $\mathbf{v}\in S^{d-1}$. Moreover, its sub-Gaussian norm is defined as follows:

\begin{equation*}
\|x\|_{\psi_2}=\sup_{\mathbf{v}\in S^{d-1}}\|\langle\mathbf{x},\mathbf{v}\rangle\|_{\psi_2}
\end{equation*}
\end{definition}

\begin{lemma}\label{SSC_all}
Let $X\in\mathbb{R}^{d\times n}$ be a matrix whose columns are sampled independently from a standard Gaussian distribution, i.e., $\mathbf{x}_{i}\sim\mathcal{N}(0,I)$. Then, for any $\epsilon>0$, there is a probability of at least $1-\delta$ that $X$ satisfies:
\begin{align*}
\lambda_{max}(XX^T)\le n+(1-2\epsilon)^{-1}\sqrt{cnd+c'n\log \frac{2}{\delta}}\\
\lambda_{min}(XX^T)\ge n-(1-2\epsilon)^{-1}\sqrt{cnd+c'n\log \frac{2}{\delta}}
\end{align*}
where $c=24e^2log\frac{3}{\epsilon}$ and $c'=24e^2$.
\end{lemma}
\begin{theorem_apd}\label{SSC_local}
Let $X\in\mathbb{R}^{d\times n}$ be a matrix whose columns are sampled independently from a standard Gaussian distribution, i.e., $\mathbf{x}_{i}\sim\mathcal{N}(0,I)$. Then, for any $k>0$, there is a probability of at least $1-\delta$ that the matrix $X$ satisfies the SSC and SSS properties with constants:
\begin{align*}
&\Lambda_{k}\le k(1+3e\sqrt{6\log\frac{en}{k}})+\textit{O}(\sqrt{nd+n\log\frac{1}{\delta}})\\
&\lambda_{k}\ge n-(n-k)(1+3e\sqrt{6\log\frac{en}{n-k}})-\Omega(\sqrt{nd+n\log\frac{1}{\delta}})
\end{align*}
\end{theorem_apd}

\begin{lemma}
Let $X\in\mathbb{R}^{d\times n}$ be a matrix with columns sampled from some sub-Gaussian distribution with sub-Gaussian norm $K$ and covariance $\Sigma$. Then, for any $\delta>0$, there is a probability of at least $1-\delta$ that each of the following statements holds:
\begin{align*}
\lambda_{max}(XX^T)\le \lambda_{max}(\Sigma)\cdot n+C_{K}\cdot\sqrt{dn}+t\sqrt{n}\\
\lambda_{min}(XX^T)\ge \lambda_{min}(\Sigma)\cdot n-C_{K}\cdot\sqrt{dn}-t\sqrt{n}
\end{align*}
where $t=\sqrt{\frac{1}{c_{K}}\log \frac{2}{\delta}}$ and $c_{K}$, $C_{K}$ are absolute constants that depend only on the sub-Gaussian norm $K$ of the distribution.
\end{lemma}
\section{Supplementary Material for Proofs of CORALS}
In this section, we provide details of the convergence theory of CORALS. We begin the description with the relationship between CORALS and momentum gradient descent. Then we show two important assumptions that constrain the outlier distribution and convergence behavior, and give details of all the convergence proof. 

\setcounter{lemma}{0}
\renewcommand{\thelemma}{\arabic{lemma}}
\begin{theorem_apd}
Suppose that the number of samples $n$ and the data dimension $d$ satisfy $d/n\to 0$, $M=\tau I$, and assume that $\mathbf{x}_i\in \mathbb{R}^d$ are generated from the standard normal distribution. Then, in the $t^{th}$ iteration of CORALS:
\[
\mathbf{b}^{s+1}= HT_k(P_{MX} \mathbf{b}^{s}+(I-P_{MX})\mathbf{y}-P_{MM}\mathbf{w}^{t})
\]
can be formally considered as:
\begin{align*}
\mathbf{b}^{s+1}
&=HT_k\left(\mathbf{b}^{s}-\nabla f_{corals}(\mathbf{b}^{s})\right)
\\
&=HT_k\left[\mathbf{b}^{s}-(A\nabla f_{crr}(\mathbf{b}^{s})+B\nabla f_{crr}(\hat{\mathbf{b}}_{t})+C)\right]
\end{align*}
where $A+B=I$, $\|C\|_2=O(\sqrt{d})$. 
\end{theorem_apd}
\begin{proof}
\begin{align*}
\mathbf{b}^{s+1}&= HT_k(P_{MX} \mathbf{b}^{s}+(I-P_{MX})\mathbf{y}-P_{MM}\mathbf{w}^{t})
\\
&=HT_k\left(\mathbf{b}^{s}+(I-P_{MX})(\mathbf{y}-\mathbf{b}^{s})-P_{MM}\mathbf{w}^{t}\right)
\\
&=HT_k\left(\mathbf{b}^{s}-\nabla f_{corals}(\mathbf{b}^{s})\right)
\end{align*}
where $\nabla f_{corals}(\mathbf{b}^{s})=(P_{MX}-I)(\mathbf{y}-\mathbf{b}^{s})+P_{MM}\mathbf{w}^{t}$. According to Theorem 5.39 in \cite{vershynin2010introduction}, there is a probability of at least $1-\delta$ that:
\begin{equation*}
\left\|\frac{1}{n}XX^{T}-I\right\|_2\le \frac{1}{2}\sqrt{\frac{d}{n}}+\frac{u}{\sqrt{n}}
\end{equation*}
where $u=\sqrt{2\log\frac{2}{\delta}}$. It can be easily deduced that the inverse of a matrix has the same properties when $d/n$ is relatively small:
\begin{equation*}
\left\|\left(\frac{1}{n}XX^{T}\right)^{-1}-I\right\|_2\le \frac{1}{2}\sqrt{\frac{d}{n}}+\frac{u}{\sqrt{n}}
\end{equation*}
Then, we obtain the following property by applying the above theory:
\begin{align*}
\left\|X^{T}(XX^{T})^{-1}X-\frac{1}{n}X^{T}X\right\|_2\le \frac{\Lambda_n}{n} \left\|\left(\frac{1}{n}XX^{T}\right)^{-1}-I\right\|_2\le \frac{1}{2}\sqrt{\frac{d}{n}}+\frac{u}{\sqrt{n}}
\end{align*}
With the same proof steps, we have:
\begin{equation*}
\left\|X^{T}(XX^{T}+M)^{-1}X-\frac{1}{n+\tau}X^{T}X\right\|_2\le \frac{1}{2}\sqrt{\frac{d}{n}}+\frac{u}{\sqrt{n}}
\end{equation*}
Noticing that $\mathbf{w}^{t}=(XX^{T})^{-1}X(y-\hat{\mathbf{b}}_t)$, we insert this information into the gradient $\nabla f_{corals}(\mathbf{b}^{s})$:
\begin{align*}
\nabla f_{corals}(\mathbf{b}^{s})&=(P_{MX}-I)(\mathbf{y}-\mathbf{b}^{s})+P_{MM}\mathbf{w}^{t}
\\
&=(P_{MX}-I)(\mathbf{y}-\mathbf{b}^{s})+P_{MM}(XX^{T})^{-1}X(y-\hat{\mathbf{b}}_t)
\\
&=(\frac{1}{n+\tau}X^{T}X-I)(\mathbf{y}-\mathbf{b}^{s})+P_{MM}(XX^{T})^{-1}X(y-\hat{\mathbf{b}}_t)+C
\\
&=(\frac{1}{n+\tau}X^{T}X-I)(\mathbf{y}-\mathbf{b}^{s})+\frac{\tau}{n(n+\tau)}X^{T}X(y-\hat{\mathbf{b}}_t)+C
\\
&=(\frac{n}{n+\tau}P_{X}-I)(\mathbf{y}-\mathbf{b}^{s})+\frac{\tau}{(n+\tau)}P_{X}(y-\hat{\mathbf{b}}_t)+C
\\
&=A\nabla f_{crr}(\mathbf{b}^{s})+B\nabla f_{crr}(\hat{\mathbf{b}}_t)+C
\end{align*}
where $A=(\frac{n}{n+\tau}P_{X}-I)(P_{X}-I)^{-1}$, $B=(\frac{\tau}{n+\tau}P_{X})(P_{X}-I)^{-1}$, and $\|C\|_2=O(\sqrt{d})$. Though the matrix $P_{X}-I$ is not actually invertible,  $A$ and $B$ are pseudo-matrices, but they still satisfy the relationship $A+B=I$.

\end{proof}

\begin{assumption}\label{assumption 1}
For any subset $S_k\subseteq [n]$, $|S_k|\le k$,  then for a constant non-negative number $\gamma$, that in any iteration of CORALS, the following statement is true:
\begin{equation*}
\|\boldsymbol{\epsilon}_{S_k}\|_2
\le
\gamma\max_{S_k\subseteq [n]}\|\mathbf{g}_{S_{k}}\|_2
\end{equation*}
\end{assumption}
The meaning of Assumption \ref{assumption 1} is that the algorithm has not yet converged. From Lemma 5 in Bhatia's work \cite{bhatia2015robust}, there is a probability of at least $1-\delta$ that the upper bound of $\|\boldsymbol{\epsilon}_{S_k}\|_2$ can be given as:
\[
\|\boldsymbol{\epsilon}_{S_k}\|_2\le \sigma \sqrt{k}\sqrt{1+2e\sqrt{6\log\frac{en}{\delta k}}},
\]
Thus, $\|\boldsymbol{\epsilon}_{S_k}\|_2$ itself is also $O(\sigma \sqrt{k})$ if $k=O(n)$. If Assumption \ref{assumption 1} is true, this indicates that in the $t^{th}$ iteration of CORALS: 
\begin{equation*}
O(\sigma \sqrt{k})
\le
\gamma\max_{S_k\in S}\|\mathbf{g}_{S_{k}}\|_2
\le\gamma\frac{\sqrt{\Lambda_k}\lambda_{max}(M)}{\lambda_{min}(XX^{T}+M)}\|\mathbf{w}^{*}-\mathbf{w}^{t}\|_2
\end{equation*}
As long as $\lambda_{max}(M)$ is $O(n)$, then $\frac{\lambda_{max}(M)}{\lambda_{min}(XX^{T}+M)}\le \frac{\lambda_{max}(M)}{\lambda_{min}(XX^{T})+\lambda_{min}(M)}=O(1)$. Note that the upper bound of $\Lambda_k$ is also $O(k)$ when $k=O(n)$, as seen in Theorem \ref{SSC_local}. This indicates that $\|\mathbf{w}^{*}-\mathbf{w}^{t}\|_2\ge \frac{1}{\gamma}O(\sigma)$. This assumption is more likely to be true when $\gamma$ is large. If the assumption is no longer valid, the algorithm has already converged to the desired result.

\begin{assumption}\label{assumption 2}
Define $I_{*}=supp(\mathbf{b}^{*})$ and $I_{\tilde{\mathbf{b}}}=supp(\tilde{\mathbf{b}})$. Then, $|I_{*}/I_{\tilde{\mathbf{b}}}|=o(n)$.
\end{assumption}
Assumption \ref{assumption 2} ensures the identifiability of $\mathbf{b}^*$. This assumption is directly derived from the definition of the robust regression problem:
\begin{equation}
(\hat{\mathbf{w}},\hat S)=\arg\min_{\substack{\mathbf{w}\in\mathbb{R}^{p},S\subset[n]\\ |S|=n-k}}\sum_{i \in S}(y_{i}-\mathbf{x}_{i}^{T}\mathbf{w})^2 
\end{equation}
Suppose that $k=k^*$ in this situation. Then, if $|I_{*}/I_{\tilde{\mathbf{b}}}|=O(n)$, we consider the estimation error on two sets:
\begin{equation}\label{assum 2}
\|\mathbf{y}_{I_{\tilde{\mathbf{b}}}^c}-X_{I_{\tilde{\mathbf{b}}}^c}^{T}\tilde{\mathbf{w}}\|_2
\le
\|\mathbf{y}_{I_{\tilde{\mathbf{b}}}^c}-X_{I_{\tilde{\mathbf{b}}}^c}^{T}\mathbf{w}^{*}\|_2
<
\|\mathbf{y}_{I_{*}^c}-X_{I_{*}^c}^{T}\mathbf{w}^{*}\|_2
\le
\|\mathbf{y}_{I_{*}^c}-X_{I_{*}^c}^{T}\hat{\mathbf{w}}\|_2+ o(n)
\end{equation}
where 
\begin{align*}
\tilde{\mathbf{w}}&=\arg\min_{\substack{\mathbf{w}\in\mathbb{R}^{d}}}\|\mathbf{y}_{I_{\tilde{\mathbf{b}}}^c}-X_{I_{\tilde{\mathbf{b}}}^c}^{T}\mathbf{w}\|_2
\\
\hat{\mathbf{w}}&=\arg\min_{\substack{\mathbf{w}\in\mathbb{R}^{d}}}\|\mathbf{y}_{I_{*}^c}-X_{I_{*}^c}^{T}\mathbf{w}\|_2
\end{align*}
The last inequality of Eq. (\ref{assum 2}) holds because the parameter $\hat{\mathbf{w}}$ converges to the true parameter $\mathbf{w}^{*}$ as $n\to \infty$. Thus, for a robust regression algorithm, it is impossible to distinguish the true uncorrupted subset, and the estimation $\tilde{\mathbf{w}}$ will have an unavoidable bias as $\mathbf{b}_{I_{*}/I_{\tilde{\mathbf{b}}}}^{*}$ cannot be removed. As a result, we need Assumption \ref{assumption 2} to ensure that the error term $\mathbf{b}^*$ is identifiable.
\setcounter{lemma}{1}
\renewcommand{\thelemma}{\arabic{lemma}}
\begin{theorem_apd}
Let $X=[\mathbf{x}_1,\dots,\mathbf{x}_n]\in \mathbb{R}^{d\times n}$ be the given data matrix and $\mathbf{y}=X^{T}\mathbf{w}^{*}+\mathbf{b}^{*}+\boldsymbol{\epsilon}$ be the corrupted output with sparse corruptions of $\|\mathbf{b}^*\|_0\le k\cdot n$. The elements of $\boldsymbol{\epsilon}$ are independent and follow the normal distribution $\mathcal{N}(0,\sigma^2)$. For a specific positive semi-definite matrix $M$, the data matrix $X$ satisfies the SSC and SSS properties such that $2\frac{\Lambda_{2k}}{\lambda_{min}(XX^T+M)}<1$. Then, in the $t^{th}$ iteration of CORALS, if $k>k^{*}$, it is guaranteed with a probability of at least $1-\delta$ that, for any $\epsilon,\delta>0$, after $S_{0}=\textit{O}(\log (\frac{\|\mathbf{b}^*\|_2}{\varepsilon}))$ iterations of TRIP:
\begin{align*}
\|\mathbf{b}_{I_{s+1}}^{s+1}-\tilde{\mathbf{b}}_{I_{s+1}}\|_2
\le
\epsilon
+
O\left(\sigma\sqrt{d\log\frac{d}{\delta}}+\sigma o(n)\right)
+
2\frac{(\sqrt{\Lambda_{2k}}+\gamma\sqrt{\Lambda_{k}})\lambda_{max}(M)}{\lambda_{min}(XX^T+M)}\|\mathbf{w}^{*}-\mathbf{w}^{t}\|_2
\end{align*}
\end{theorem_apd}
\begin{proof}
In the $t^{th}$ iteration of CORALS, each substep of TRIP can be simplified to:
\[
\mathbf{b}^{s+1}=HT_{k}(\mathbf{b}^{*}+\boldsymbol{\epsilon}+P_{MX}(\mathbf{b}^{s}-\mathbf{b}^{*}-\boldsymbol{\epsilon})+\mathbf{g})
\]
where $\mathbf{g}=P_{MM}(\mathbf{w}^{*}-\mathbf{w}^{t})$. We define $\tilde{\mathbf{b}}=HT_{k}(\mathbf{b}^{*}+\boldsymbol{\epsilon})$ and $I_{s+1}=supp(\mathbf{b}^{s+1})\cup supp(\tilde{\mathbf{b}})$. From the properties of the hard thresholding operator, we obtain the following inequality:
\begin{align*}
&\|\mathbf{b}_{I_{s+1}}^{s+1}-(\mathbf{b}_{I_{s+1}}^{*}+\boldsymbol{\epsilon}_{I_{s+1}}+X_{I_{s+1}}^{T}(XX^{T}+M)^{-1}X(\mathbf{b}^{s}-\mathbf{b}^{*}-\boldsymbol{\epsilon})+\mathbf{g}_{I_{s+1}})\|_2
\\
&\le\|\tilde{\mathbf{b}}_{I_{s+1}}-(\mathbf{b}_{I_{s+1}}^{*}+\boldsymbol{\epsilon}_{I_{t+1}}+X_{I_{s+1}}^{T}(XX^{T}+M)^{-1}X(\mathbf{b}^{s}-\mathbf{b}^{*}-\boldsymbol{\epsilon})+\mathbf{g}_{I_{s+1}})\|_2
\end{align*}
By incorporating $\tilde{\mathbf{b}}$ into the above inequality and using the trigonometric inequality, the error term can be divided into four terms:
\begin{align*}
&\|\mathbf{b}_{I_{s+1}}^{s+1}-\tilde{\mathbf{b}}_{I_{s+1}}\|_2
\\ 
&\le2\|\tilde{\mathbf{b}}_{I_{s+1}}-(\mathbf{b}_{I_{s+1}}^{*}+\boldsymbol{\epsilon}_{I_{s+1}}+X_{I_{s+1}}^{T}(XX^{T}+M)^{-1}X(\mathbf{b}^{s}-\mathbf{b}^{*}-\boldsymbol{\epsilon})+\mathbf{g}_{I_{s+1}})\|_2
\\
&\le
2\|X_{I_{s+1}}^{T}(XX^{T}+M)^{-1}X(\mathbf{b}^{s}-\mathbf{b}^{*}-\boldsymbol{\epsilon})\|_2+2\|\tilde{\mathbf{b}}_{I_{s+1}}-(\mathbf{b}_{I_{s+1}}^{*}+\boldsymbol{\epsilon}_{I_{s+1}})\|_2+2\|\mathbf{g}_{I_{s+1}}\|_2
\\
&\le
2\|X_{I_{s+1}}^{T}(XX^{T}+M)^{-1}X(\mathbf{b}^{s}-\tilde{\mathbf{b}})\|_2
+2\underbrace{\|\tilde{\mathbf{b}}_{I_{s+1}}-(\mathbf{b}_{I_{s+1}}^{*}+\boldsymbol{\epsilon}_{I_{s+1}})\|_2}_{\normalsize{\textcircled{\scriptsize{1}}}\normalsize}
\\
&\quad +2\underbrace{\|X_{I_{s+1}}^{T}(XX^{T}+M)^{-1}X(\tilde{\mathbf{b}}-\mathbf{b}^{*}-\boldsymbol{\epsilon})\|_2}_{\normalsize{\textcircled{\scriptsize{2}}}\normalsize}+2\|\mathbf{g}_{I_{s+1}}\|_2
\end{align*}
We first consider term $\normalsize{\textcircled{\scriptsize{1}}}\normalsize$. By applying the properties of the hard thresholding operator, we have:
\begin{align*}
\normalsize{\textcircled{\scriptsize{1}}}\normalsize=\|\tilde{\mathbf{b}}_{I_{s+1}}-(\mathbf{b}_{I_{s+1}}^{*}+\boldsymbol{\epsilon}_{I_{s+1}})\|_2
&\le
\|\boldsymbol{\epsilon}_{I_{s+1}/I_{\tilde{\mathbf{b}}}}\|_2
\end{align*}
Through Assumption \ref{assumption 1}, we can specify an upper bound of term $\normalsize{\textcircled{\scriptsize{1}}}\normalsize$ as: 
\begin{align*}
\|\boldsymbol{\epsilon}_{I_{s+1}/I_{\tilde{\mathbf{b}}}}\|_2
&\le\gamma\max_{S_{k}\subseteq [n]}\|g_{S_{k}}\|_2
\\
&=\gamma \max_{S_{k}\subseteq [n]}\left\|X^{T}_{S_{k}}(XX^{T}+M)^{-1}M(\mathbf{w}^{*}-\mathbf{w}^{t}) \right\|_2
\\
&\le\gamma\frac{\sqrt{\Lambda_{k}}\lambda_{max}(M)}{\lambda_{min}(XX^T+M)}\|\mathbf{w}^*-\mathbf{w}^t\|_2
\end{align*}
As for term $\normalsize{\textcircled{\scriptsize{2}}}\normalsize$, we find that:
\begin{align*}
\normalsize{\textcircled{\scriptsize{2}}}\normalsize=
\|X_{I_{s+1}}^{T}(XX^{T}+M)^{-1}X(\tilde{\mathbf{b}}-\mathbf{b}^{*}-\boldsymbol{\epsilon})\|_2
\le
\frac{\sqrt{\Lambda_{2k}}}{\lambda_{min}(XX^T+M)}\|X(\tilde{\mathbf{b}}-\mathbf{b}^{*}-\boldsymbol{\epsilon})\|_2
\end{align*}
According to Assumption \ref{assumption 2}, the upper bound of $\normalsize{\textcircled{\scriptsize{2}}}\normalsize$ can be written as:
\begin{align*}
\frac{\sqrt{\Lambda_{2k}}}{\lambda_{min}(XX^T+M)}&\|X(\tilde{\mathbf{b}}-\mathbf{b}^{*}-\boldsymbol{\epsilon})\|_2
\\
&\le 
\frac{\sqrt{\Lambda_{2k}}}{\lambda_{min}(XX^T+M)}\left(\|X\boldsymbol{\epsilon}\|_2+\|X(\tilde{\mathbf{b}}-\mathbf{b}^{*})\|_2 \right)
\\
&\le
\frac{\sqrt{\Lambda_{2k}}}{\lambda_{min}(XX^T+M)}\left(O\left(\sigma\sqrt{nd\log\frac{d}{\delta}}\right)+\|X(\tilde{\mathbf{b}}-\mathbf{b}^{*})\|_2\right)
\\
&\le
\frac{\sqrt{\Lambda_{2k}}}{\lambda_{min}(XX^T+M)}\left(O\left(\sigma\sqrt{nd\log\frac{d}{\delta}}\right)+\sqrt{\Lambda_n}\max_{S_{o(n)}\subseteq [n]}\|\boldsymbol{\epsilon}_{S_{o(n)}}\|_2\right)
\\
&\le
\frac{\sqrt{\Lambda_{2k}}}{\lambda_{min}(XX^T+M)}\left(O\left(\sigma\sqrt{nd\log\frac{d}{\delta}}\right)+\sqrt{\Lambda_n}\sigma o(\sqrt{n})\right)
\\
&\le
O\left(\sigma\sqrt{d\log\frac{d}{\delta}}+\sigma o(\sqrt{n})\right)
\end{align*}
The second inequality above can be found in Bhatia et al.~\cite{bhatia2017consistent}, and the last inequality comes from $\sqrt{\Lambda_{2k}}=O(\sqrt{n})$ when $k=O(n)$. The other two terms in $\|\mathbf{b}_{I_{s+1}}^{s+1}-\tilde{\mathbf{b}}_{I_{s+1}}\|_2$ can be easily calculated by:
\begin{align*}
\|X_{I_{s+1}}^{T}(XX^{T}+M)^{-1}X(\mathbf{b}^{s}-\tilde{\mathbf{b}})\|_2
&=
\|X_{I_{s+1}}^{T}(XX^{T}+M)^{-1}X_{I_{s}}(\mathbf{b}^{s}-\tilde{\mathbf{b}})\|_2
\\
&\le
\frac{\Lambda_{2k}}{\lambda_{min}(XX^T+M)}\|\mathbf{b}^{s}-\tilde{\mathbf{b}}\|_2
\end{align*}
\begin{align*}
\|\mathbf{g}_{I^{s+1}}\|_2&=\|X_{I^{s+1}}^T(XX^T+M)^{-1}M(\mathbf{w}^*-\mathbf{w}^t)\|_2\\
&\le\frac{\sqrt{\Lambda_{2k}}\lambda_{max}(M)}{\lambda_{min}(XX^T+M)}\|\mathbf{w}^*-\mathbf{w}^t\|_2
\end{align*}
As a result, the error bound of $\|\mathbf{b}_{I_{s+1}}^{s+1}-\tilde{\mathbf{b}}_{I_{s+1}}\|_2$ can be set as:
\begin{align*}
\|\mathbf{b}_{I_{s+1}}^{s+1}-\tilde{\mathbf{b}}_{I_{s+1}}\|_2
\le&
2\frac{\Lambda_{2k}}{\lambda_{min}(XX^T+M)}\|\mathbf{b}^{s}-\tilde{\mathbf{b}}\|_2
+O\left(\sigma\sqrt{d\log\frac{d}{\delta}}+\sigma o(\sqrt{n})\right)
\\
&+
2\frac{(\sqrt{\Lambda_{2k}}+\gamma\sqrt{\Lambda_{k}})\lambda_{max}(M)}{\lambda_{min}(XX^T+M)}\|\mathbf{w}^{*}-\mathbf{w}^{t}\|_2
\end{align*}
Thus, as long as $2\frac{\Lambda_{2k}}{\lambda_{min}(XX^T+M)}<1$, after $S_{0}=\textit{O}(\log (\frac{\|\mathbf{b}^*\|_2}{\varepsilon}))$ iterations, we establish the following convergence property of $\mathbf{b}^{s}$:
\begin{align*}
\|\mathbf{b}_{I_{s+1}}^{s+1}-\tilde{\mathbf{b}}_{I_{s+1}}\|_2
\le
\epsilon
+
O\left(\sigma\sqrt{d\log\frac{d}{\delta}}+\sigma o(\sqrt{n})\right)
+
2\frac{(\sqrt{\Lambda_{2k}}+\gamma\sqrt{\Lambda_{k}})\lambda_{max}(M)}{\lambda_{min}(XX^T+M)}\|\mathbf{w}^{*}-\mathbf{w}^{t}\|_2
\end{align*}
\end{proof}

\begin{theorem_apd}
Under the conditions of Theorem 2 and assuming that $\mathbf{x}_i\in \mathbb{R}^d$ are generated from the standard normal distribution, for $k>k^{*}$, it is guaranteed with a probability of at least $1-\delta$ that, if:
\begin{equation*}
2\frac{\sqrt{\Lambda_n}(\sqrt{\Lambda_{2k}}+\gamma\sqrt{\Lambda_{k}})\lambda_{max}(M)}{\lambda_n\lambda_{min}(XX^T+M)}<1
\end{equation*}
then, for any $\epsilon,\delta>0$, after $T_{0}=O(log(\frac{\|\mathbf{w}^{0}-\mathbf{w}^{*}\|_2}{\epsilon}))$ iterations of CORALS, the current estimation coefficient $\mathbf{w}^{t}$ will satisfy: 
\[\|\mathbf{w}^{t}-\mathbf{w}^*\|_2
\le
\epsilon
+
O(\sigma\sqrt{\frac{d}{n}\log \frac{d}{\delta}})
+
\frac{1}{\gamma}O(\sigma)
\]
\end{theorem_apd}
\begin{proof}
In the $t^{th}$ iteration of CORALS:
\begin{align*}
\mathbf{w}^{t+1}
&=(XX^T)^{-1}X(\mathbf{y}-\mathbf{b}^s)
=(XX^T)^{-1}X(X^T\mathbf{w}^*+\mathbf{b}^*+\boldsymbol{\epsilon}-\mathbf{b}^s)
\\
&=\mathbf{w}^*+(XX^T)^{-1}X(\boldsymbol{\epsilon}+\mathbf{b}^*-\tilde{\mathbf{b}}+\tilde{\mathbf{b}}-\mathbf{b}^s)
\\
&=\mathbf{w}^*+(XX^T)^{-1}X(\boldsymbol{\epsilon}+\mathbf{b}^*-\tilde{\mathbf{b}})+(XX^T)^{-1}X(\tilde{\mathbf{b}}-\mathbf{b}^s)
\end{align*}
Thus:
\begin{align*}
\|\mathbf{w}^{t+1}-\mathbf{w}^*\|_2
&\le 
\|(XX^T)^{-1}X(\boldsymbol{\epsilon}+\mathbf{b}^*-\tilde{\mathbf{b}})\|+2
+ \|(XX^T)^{-1}X(\tilde{\mathbf{b}}-\mathbf{b}^s)\|_2
\\
&\le
\frac{1}{\lambda_n}\|X(\boldsymbol{\epsilon}+\mathbf{b}^*-\tilde{\mathbf{b}})\|_2
+
\frac{\sqrt{\Lambda_n}}{\lambda_n}\|\tilde{\mathbf{b}}-\mathbf{b}^s\|_2
\\
&\le
\frac{1}{\lambda_n}O\left(\sigma\sqrt{nd\log\frac{d}{\delta}}+\sqrt{\Lambda_n}\sigma o(\sqrt{n})\right)
\\
&\quad +2\frac{\sqrt{\Lambda_n}}{\lambda_n} 
\left(\epsilon
+
O\left(\sigma\sqrt{d\log\frac{d}{\delta}}+\sigma o(\sqrt{n})\right)
+
\frac{(\sqrt{\Lambda_{2k}}+\gamma\sqrt{\Lambda_{k}})\lambda_{max}(M)}{\lambda_{min}(XX^T+M)}\|\mathbf{w}^{*}-\mathbf{w}^t\|_2
\right)
\\
&= O(\sigma\sqrt{\frac{d}{n}\log \frac{d}{\delta}})
+
2\frac{\sqrt{\Lambda_n}(\sqrt{\Lambda_{2k}}+\gamma\sqrt{\Lambda_{k}})\lambda_{max}(M)}{\lambda_n\lambda_{min}(XX^T+M)}\|\mathbf{w}^{*}-\mathbf{w}^t\|_2
\end{align*}
Then, if: 
\begin{equation*}
2\frac{\sqrt{\Lambda_n}(\sqrt{\Lambda_{2k}}+\gamma\sqrt{\Lambda_{k}})\lambda_{max}(M)}{\lambda_n\lambda_{min}(XX^T+M)}<1
\end{equation*}
after $T_{0}=O(log(\frac{\|\mathbf{w}^{0}-\mathbf{w}^{*}\|_2}{\epsilon}))$ iterations, and we include the error from applying Assumption \ref{assumption 1}, then we obtain the final estimation error of CORALS as:
\[\|\mathbf{w}^{t}-\mathbf{w}^*\|_2
\le
\epsilon
+
O(\sigma\sqrt{\frac{d}{n}\log \frac{d}{\delta}})
+
\frac{1}{\gamma}O(\sigma)
\]
\end{proof}

\begin{theorem_apd}
Suppose that $\mathbf{x}_i\in \mathbb{R}^d$ are generated from the standard normal distribution. For $k>k^{*}$, it is guaranteed with a probability of at least $1-\delta$ that, if $M$ is in the form $\tau I$, and $\gamma=1$, the maximum breakdown point of CORALS can reach up to $1\%$ when $\tau=0.049n$.
\end{theorem_apd}
\begin{proof}
Our purpose is to find the $\tau$ that maximizes the breakdown point of CORALS under the constraint that the conditions in Theorem 1 and 2 are satisfied: 
\begin{align*}
&\max_{\tau\in \mathbb{R}_{+}, k\in [0,n]} k
\\
\begin{split}
s.t.\quad
&\left \{
\begin{array}{ll}
2\frac{\Lambda_{2k}}{\lambda_{n}+\tau}<1\\
2\tau\frac{\sqrt{\Lambda_n}(\sqrt{\Lambda_{2k}}+\sqrt{\Lambda_{k}})}{\lambda_n(\lambda_n+\tau)}<1 
\end{array} 
\right. 
\end{split} 
\end{align*}
Using the results of Theorems \ref{SSC_all} and \ref{SSC_local}, we can convert the original condition into the computable inequality: 
\begin{align*}
&\max_{\tau\in \mathbb{R}_{+}, k\in [0,n]} k
\\
\begin{split}
s.t.\quad
&\left \{
\begin{array}{ll}
\frac{4}{n+\tau}k(1+3e\sqrt{6\log\frac{en}{2k}})<1\\
\frac{2\tau}{\sqrt{n}(n+\tau)}(\sqrt{2k(1+3e\sqrt{6\log\frac{en}{2k}})}+\sqrt{k(1+3e\sqrt{6\log\frac{en}{k}})})<1 
\end{array} 
\right. 
\end{split} 
\end{align*}
This optimization problem can be solved by a two-dimensional grid search method. The final result is that the maximum breakdown point is $1\%$ when $\tau=0.049n$.
\end{proof}

\end{document}